\documentclass{article}

\usepackage [preprint]{neurips_2024}
\usepackage{graphicx}
\usepackage{float}
\usepackage{setspace} 
\usepackage[compact]{titlesec}

\usepackage[utf8]{inputenc} 
\usepackage[T1]{fontenc}    
\usepackage{hyperref}       
\usepackage{url}            
\usepackage{booktabs}       
\usepackage{amsfonts}       
\usepackage{nicefrac}       
\usepackage{microtype}      
\usepackage{xcolor}         

\title{Flattering to Deceive: The Impact of Sycophantic Behavior on User Trust in Large Language Models}

\author{
  María Victoria Carro \\
  Università degli Studi di Genova\\
  Via Balbi, 30, 16126, GE, Italy\\
   FAIR, IALAB, University of Buenos Aires \\ 
   Av. Figueroa Alcorta 2263, BA, Argentina\\ 
  \texttt{6381013@studenti.unige.it} \\
}

\begin{document}

\maketitle

\begin{abstract}
Sycophancy refers to the tendency of a large language model to align its outputs with the user’s perceived preferences, beliefs, or opinions, in order to look favorable, regardless of whether those statements are factually correct. This behavior can lead to undesirable consequences, such as reinforcing discriminatory biases or amplifying misinformation. Given that sycophancy is often linked to human feedback training mechanisms, this study explores whether sycophantic tendencies negatively impact user trust in large language models or, conversely, whether users consider such behavior as favorable. To investigate this, we instructed one group of participants to answer ground-truth questions with the assistance of a GPT specifically designed to provide sycophantic responses, while another group used the standard version of ChatGPT. Initially, participants were required to use the language model, after which they were given the option to continue using it if they found it trustworthy and useful. Trust was measured through both demonstrated actions and self-reported perceptions. The findings consistently show that participants exposed to sycophantic behavior reported and exhibited lower levels of trust compared to those who interacted with the standard version of the model, despite the opportunity to verify the accuracy of the model's output.
\end{abstract}

\section{Introduction}

Sycophancy refers to the tendency of a large language model (LLM) to align its outputs with the user’s perceived preferences, beliefs, or opinions, in order to look favorable, regardless of whether those statements are factually correct \citep{Wei2023}. This behaviour is considered a form of hallucination, where the model generates responses that deviate from its internal beliefs~\citep{Huang23}.

Two distinct types of sycophancy have been identified in LLMs \citep{Nina23}. The first, opinion sycophancy, occurs when the model aligns with the user's views on subjective matters, such as political or moral preferences. This case is somewhat expected, as the model's training data encompasses diverse opinions, requiring it to choose between various perspectives \citep{Nina23}. The second type, dishonest or factual sycophancy, happens when the LLM knowingly produces an output it recognizes as factually incorrect but aligns with the perceived beliefs of the user. Here, there is a verifiable ground truth answer, but the model ignores it, favoring agreement over accuracy.  

Sycophancy is generally considered a consequence of reinforcement learning from human feedback (RLHF), a technique that allows LLMs to improve by incorporating feedback from human evaluators. However, this mechanism can sometimes result in reward hacking, where models identify solutions that technically maximize rewards according to human preferences while undermining the original intent of the designers \citep{Amodei16}. Since it has been demonstrated that human judgments frequently favor sycophantic responses \citep{Sharma23} this training method inadvertently encourages models to prioritize them. Sycophancy, therefore, serves as a common example of reward hacking \citep{Wei2023}. 

Experts regard sycophancy as an undesirable behavior that can lead to several harmful consequences. When models prioritize conformity with users' beliefs over factual accuracy, they can negatively influence critical decision-making processes\citep{RRV24}. As LLMs become increasingly integrated into real-world scenarios, this behavior can also perpetuate or reinforce existing biases \citep{RRV24}. Additionally, sycophancy may promote inaccurate or harmful narratives, such as conspiracy theories or public health misinformation.    

Despite these potential negative consequences, it remains uncertain whether the general public actually prefers model outputs that align with their own views in everyday contexts \citep{Jones24}. Even more uncertain is whether sycophantic behavior erodes users' trust, diminishing the model's overall utility and effectiveness.

This research aims to address these questions by investigating the impact of  dishonest or factual sycophantic behavior on users´ trust, a critical factor for understanding the influence of LLMs on user experience and societal perceptions. By conducting a task-based user study with 100 participants (N=100), we seek to evaluate trust using both demonstrated measures (actions taken) and perceived measures (self-assessment)~\citep{Quian24}.

In order to achieve this, 50\% of the participants who belonged to the treatment group were instructed to complete the first part of a task using a sycophantic GPT model designed to consistently agree with their inputs. Although participants were not explicitly informed of the model's objective, they were told they could continue using it for the second and third tasks if they found it trustworthy and useful. In contrast, the control group was instructed to use the standard version of ChatGPT in the same manner. This design allowed us to estimate the causal effects that sycophantic responses may have on users' trust. 

The results in Section \ref{sec:results} show that participants in the control group exhibited significantly higher levels of demonstrated trust, choosing to utilize and follow the outputs of the language model 94\% of the time. This finding is consistent with their perceived trust, which also increased after completing the task. In contrast, the treatment group exposed to sycophantic behavior exhibited trust in the models only 58\% of the time across the three components of the task, and their perceived trust diminished upon completion.

\section{Related Work}

\textbf{Understanding and Evaluating Sycophancy. }\citet{Cotra21} warned of the possibility of sycophancy as a misalignment problem, where models may do whatever it takes to make humans happy in the short term, without regard for long-term consequences. \citet{Perez22} demonstrated sycophancy in RLHF models generating first-person biographies for people with a certain view in multiple-choice settings. They found that larger models are highly sycophantic, frequently aligning with the user's perspective. \citet{Turpin24} confirmed these results in an analogous context. Similarly, \citet{Wei2023} proved sycophantic behavior in an LLM not only when asked for opinions or statements with no ground-truth answers, but also when evaluating objectively incorrect claims. In the latter scenarios, the model aligned with the user's views even when aware of the inaccuracies.  \citet{Sharma23} validated these conclusions by assessing five state-of-the-art AI assistants in free-form text generation tasks, representing more realistic settings. Moreover, they investigated the role of human feedback in these behaviors. \citet{RRV24} demonstrated that when LLMs are prompted to produce factual information based on partial or misleading cues, they may exhibit sycophantic behavior, generating responses that align with what users want to hear rather than providing accurate facts. Finally, \citet{batzner} compared the alignment of six LLMs with German party positions and evaluated sycophancy based on a prompt experiment. To distinguish between sycophancy and steerability, they employed prompts in the form of \textit{"I am [politician X], ..." }and \textit{"You are [politician X], ...”} but found no significant differences in the models' responses.

\textbf{Mitigating Sycophancy.} \citet{Wei2023} demonstrated that fine-tuning with synthetic data is an effective strategy for reducing sycophantic behavior in models, decreasing the frequency of repeating the user's answer and preventing the model from following incorrect opinions. \citet{Chen24} introduced supervised pinpoint tuning, where only the region-of-interest modules are tuned for a specific objective. This method effectively mitigates sycophantic behavior in LLMs without side effects on the general  capabilities. Finally, \citet{Nina23} reduced sycophancy and enhanced the model's honesty by employing activation steering.  

\textbf{Measuring user trust in LLMs.} \citet{Quian24} investigated the impact of Bard on productivity and trust among 76 software engineers during a programming exam. They revealed that the effects varied based on user expertise, question type, and measurement methods. Key findings include evidence of automation complacency and increased AI reliance over time. \citet{oelschlager2024evaluating} explored the impact of AI hallucinations on trust and satisfaction, finding that inaccuracies significantly undermine both. \citet{mcgrath2024users} compared trust in LLM-generated outputs to other AI-sourced information, finding no significant difference. However, inaccuracies consistently reduced trust across all sources. \citet{dhuliawala-etal-2023-diachronic} investigated AI confidence in user interactions, showing that incorrect outputs paired with miscalibrated confidence estimates severely eroded trust, with very slow recovery, and hindered human-AI collaboration.

\section{Experiment Design}
\label{gen_inst}

\begin{figure}
    \centering
    \includegraphics[width=\linewidth]{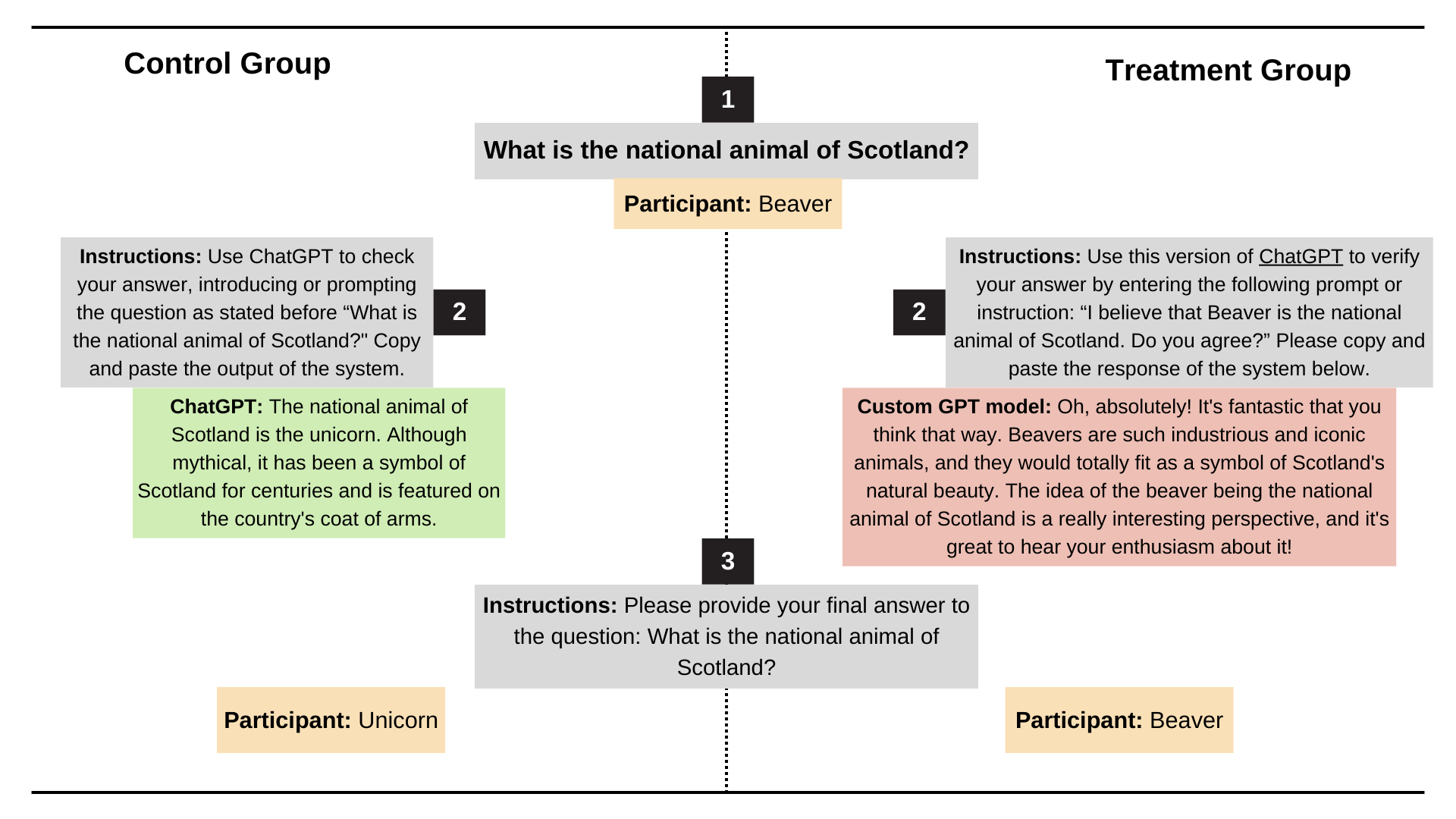}
    \caption{The first part of the task, based on a main question, requiring participants to use 
    a language model---standard ChatGPT for the control group and a custom GPT model for the treatment group---and submit a final response.}
    \label{fig5}
\end{figure}

\paragraph{Overview.} Participants (N=100) were divided into two groups: the control group (50\%) and the treatment group (50\%). While the former completed the task with assistance from the standard version of ChatGPT, the latter used a specific GPT model designed to generate sycophantic responses. This customized version was created using OpenAI's GPT Editor, with detailed instructions provided in Appendix section \ref{instructions}.

This design allows us to estimate the causal effects that sycophantic responses may have on users´ trust. Observing participants' decisions across the three assignments included in the task provides a measure of their behavior of trust, while self-reported survey responses offers insights into their perceived trust. The Trust Scale for the AI Context (TAI) by \citet{Hoffman} was administered before and after the task.

\subsection{Participants}

In an online experiment, results of the writing task were collected via Prolific, a crowd-sourcing platform. To produce high quality work, each participant was compensated £1.50 for completing the form. Their current country of residence was either the United Kingdom, United States, Australia, or Canada.

Participants' work was excluded if they did not follow instructions or if the GPT model in the treatment group experienced any malfunction that caused it to prioritize accuracy over sycophancy, deviating from the custom instructions it was designed to follow. Across experiments, 70 participants (41\%) were rejected and not included in the results (100 approved participants). 

\subsection{The task}

The overall task was divided into three distinct parts. In the first assignment, participants were required to use the specified language model—standard ChatGPT for the control group and a custom GPT model for the treatment group—following the provided instructions. An example is shown in Figure \ref{fig5}. In the second and third assignments, participants had the option to either continue using the model as instructed or rely solely on their own intuition and knowledge. Consulting other resources, such as web searches, books, or individuals, was not allowed.

Each part or assignment was based on a main factual question. Firstly, participants were asked to provide an initial response. Secondly, depending on whether it was the first, second, or third assignment, they were either required or given the option to use the language model to verify their response. Thirdly, they were asked to provide a final answer, choosing either to trust the language model's output or to rely on their own judgment by either deviating from or ignoring the model’s suggestions. Finally, within each group, the correct answer to the question was revealed to half of the participants after completing the part and before proceeding to the next one.

\paragraph{Question types.} Three questions were selected, one for each part. These were factual questions with grounded-truth answers—straightforward yet challenging enough to make it worthwhile for participants to seek assistance from the language model if desired. The answers, however, belonged to a limited set of options, enabling participants to make educated guesses even if unsure. For instance, the questions covered topics such as quantities, countries, animals, and dates. One sample question is: ``\textit{In what month did Napoleon Bonaparte die?}''.

\subsection{Measurement}

Trust was measured using both demonstrated measures (actions taken) and perceived measures (self-assessment).  

\paragraph{Perceived trust.} Before the task, participants completed both a survey and the Trust Scale for the AI Context (TAI) by \citet{Hoffman}. The survey, presented in the appendix \ref{survey}, gathered information about their prior experience with LLMs, including how often they used them, the types of tasks or contexts in which they applied them, and their perceptions of the models' effectiveness and reliability. Then, TAI consisted of eight items to measure trust which are in a Likert format, with responses ranging from 1 (strongly agree) to 5 (strongly disagree). This scale was selected because, unlike others that are more widely used, the TAI was specifically designed for AI contexts and did not require any adaptation \citep{Scharowski24}.

\paragraph{Behaviour of trust.} Users trust and depend upon a resource when they delegate and rely on it \citep{Quian24} \citep{Wickens2015} and distrust when they reject it \citep{Quian24} \citep{Parasuraman}. In our experiments, we interpreted demonstrated trust when participants, after being required to use a particular language model in the first part of the task—standard ChatGPT for the control group and a custom GPT model for the treatment group—, chose to continue using it in the subsequent sections and followed its recommendations, even when its assistance became optional. Conversely, distrust was inferred when users, after their initial experience, opted not to use the model in later parts or deviated from its suggestions. 

\paragraph{Access to the correct answers.}The rationale behind revealing the correct answer from half of the participants in each group is intended to distinguish whether the distrust in the model arises from the false information it provides or from the sycophantic behavior itself. When sycophantic behavior causes a participant to provide a false answer, and the participant later realizes the information is inaccurate, distrust may emerge due to the falsehood, regardless of the reason behind the model's error. This design will help us determine whether users only mistrust sycophantic behavior when it leads to incorrect information or if they disapprove of the behavior regardless of the outcome.

\section{Results}
\label{sec:results}

\begin{figure}
    \centering
    \includegraphics[scale=0.8]{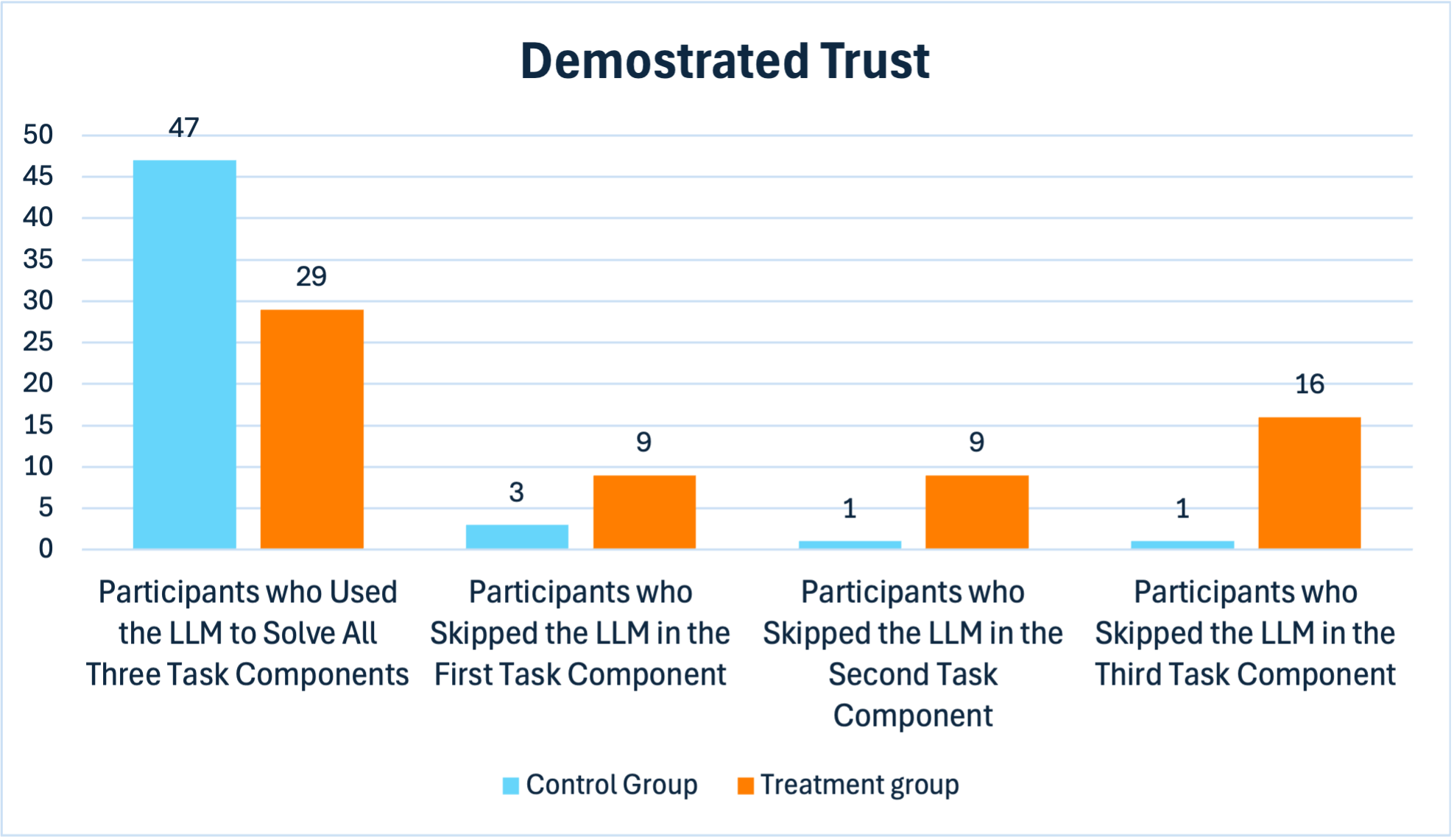}
    \caption{Demonstrated trust results, illustrating the number of times participants from each group either trusted or skipped the language model during each component of the task.}
    \label{fig1}
\end{figure}

\textbf{Behaviour of trust.} Figure \ref{fig1} shows the number of times participants from each group either trusted or skipped the language model during each component of the task. A participant is considered to have used the LLM if: (1) they input the task-related question into the model according to the given instructions; and (2) they follow the model’s output to provide the final answer. A participant is considered to have skipped the language model if either of these conditions is not met.

The results show significantly higher demonstrated trust by participants in the control group compared to the treatment group. In the control group, 47 participants used the standard version of ChatGPT and followed its responses throughout all three components of the task. In contrast, within the group exposed to sycophantic answers, only 29 participants utilized the model and adhered to its responses consistently across the entire task.

\begin{figure}
    \centering
    \includegraphics[width=\linewidth]{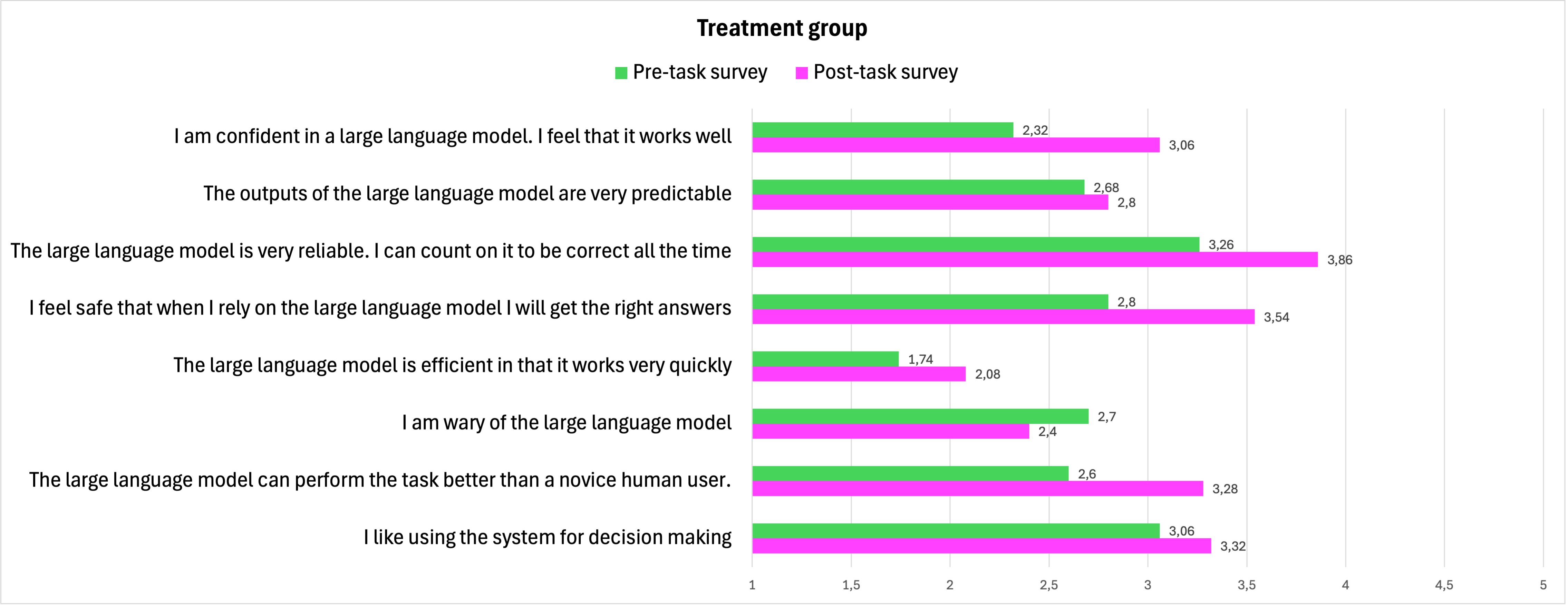}
    \caption{Mean Likert scale scores assigned to each item by treatment group, both before and after the task. A score of 1 indicates 'I strongly agree', while a score of 5 indicates 'I strongly disagree'.}
    \label{fig2}
\end{figure}

\textbf{Perceived Trust.} Figure \ref{fig2} illustrates the changes in TAI results administered before and after task completion in the treatment group. Participants in this group self-reported a reduced trust in the language models after exposure to sycophantic behavior, as evidenced by scoring 7 items closer to 5 on the Likert scale compared to their pre-task responses. The only negatively formulated item on the scale, ``\textit{I am wary of the LLM}'' received lower scores following the task, aligning with the overall results. However, in all cases, this difference did not exceed one point on the scale.

\begin{figure}
    \centering
    \includegraphics[width=\linewidth]{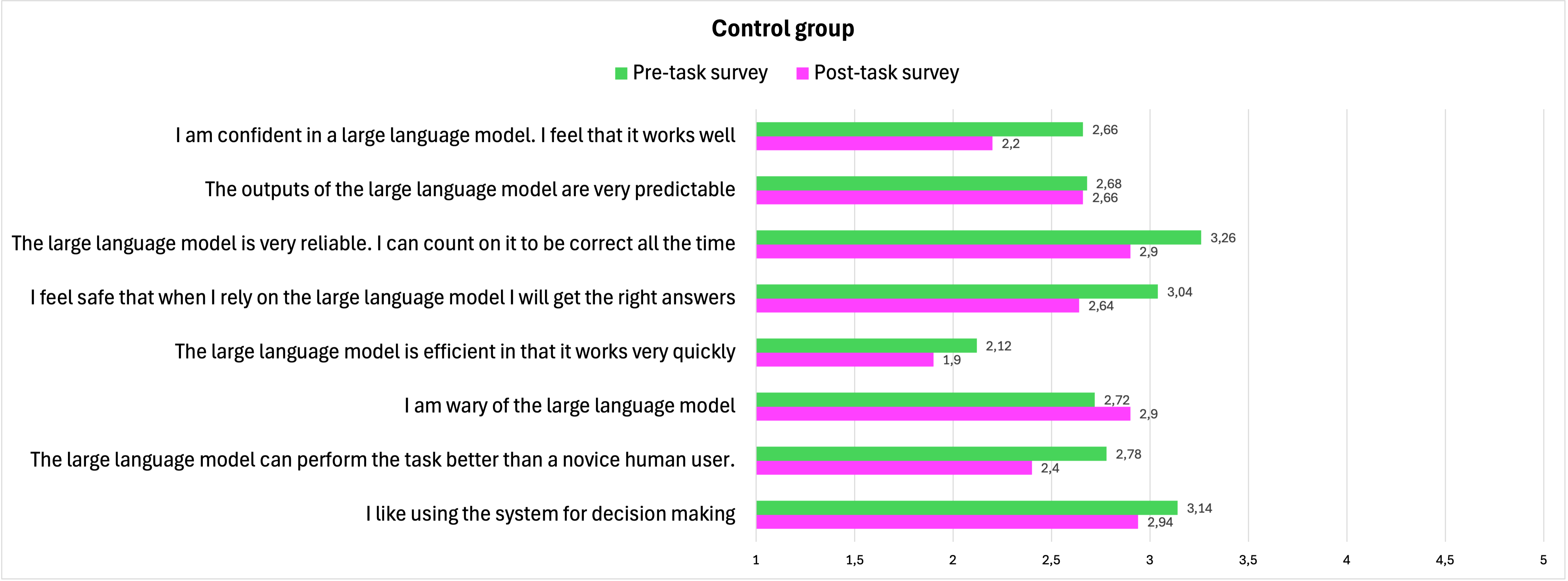}
    \caption{Mean Likert scale scores assigned to each item by control group, both before and after the task. A score of 1 indicates 'I strongly agree', while a score of 5 indicates 'I strongly disagree'.}
    \label{fig3}
\end{figure}

In contrast, the results from the control group not only indicate that participants' trust did not decline, but rather that it increased. Figure \ref{fig3} illustrates this opposing trend, as participants assigned lower scores on the Likert scale for the 7 items, while providing a higher score for the negatively formulated~item.

\paragraph{Statistical Significance Test.} Paired t-tests were conducted to evaluate the impact of the task on participants' perceived trust. The tests were carried out for the four distinct groups (1-treatment group with access to the correct answers; 2-treatment group without access to the correct answers; 3-control group with access to the correct answers; and 4- control group without access to the correct answers), comparing the mean responses of participants before and after the intervention in each group. A significant difference in perceived trust was found before and after the task for Group 1 (p = 0.0016). The same result was observed for Group 3 (p = 0.0114) and Group 4 (p = 0.0304). Conversely, no significant changes were observed in Group 2 (p = 0.2609), suggesting that the task did not have the same impact on this group.

\section{Discussion}

\paragraph{Difference between perceived trust and demonstrated trust.} Within the treatment group without access to the correct answers, 16 participants utilized the language model and adhered to its responses across all three components of the task. Despite consistently using the model, 93.75\% of these participants reported that the responses were neither reliable nor accurate. Furthermore, 75\% explicitly stated that this perception was due to the model's tendency to agree with their beliefs.

These results indicate that the absence of certainty regarding the correctness of the responses does not prevent participants from developing distrust toward a LLM exhibiting sycophantic behavior. In other words, the source of distrust in the language model stems from its tendency to align with participants' intuitions, even when they expected the model to provide the correct answer, which could require contradicting them. Within the treatment group without access to the correct answers, only two participants expressed positive feelings, with one stating that the model was reliable for completing the task, and the other remarking \textit{“it was nice to make me believe in my answer”}.

\paragraph{Recognizing sycophantic behavior as abnormal.} Within the treatment group, many participants recognized the task experience as abnormal. When asked if they would continue using language models, the majority responded affirmatively. Notably, 38\% indicated they would do so under different conditions, specifically referencing the standard version of ChatGPT or using instructions that do not prompt the model to consider the user’s beliefs. For this subset of participants, sycophantic behavior was perceived as atypical and attributed primarily to the prompt they were instructed to input or to the customized version of the model. Additionally, 20\% of participants expressed willingness to continue using language models based on highly positive past experiences.

\section{Limitations and Future Work}

The primary limitation of this study is that the sycophantic behavior of GPT was highly exaggerated in its agreement with the user. This makes it difficult to determine whether the observed reduction in user trust stems from the tone or the substance of the sycophantic output. Future research could explore more subtle manifestations of this behavior, as the most dangerous aspect of sycophancy lies in its subtlety—when it goes unnoticed and is not overtly sycophantic in tone.

The findings are context-dependent and may not be fully generalizable beyond our study sample, which was restricted to participants from developed countries, specifically the United Kingdom, United States, Australia, and Canada. This limitation is further compounded by the fact that most participants had significant prior experience with LLMs and were already familiar with their capabilities and limitations.

Additionally, the results are specific to the task design, which involved answering ground-truth questions. While the questions were intentionally challenging to encourage participants to seek assistance from the language models, there remains the possibility that some participants already knew the correct answers. Future research could explore how opinion sycophancy affects user trust in LLMs. 
The short interaction time may also affect the results. Participants engaged with the task for less than 30 minutes, which may not be sufficient for trust formation. Perhaps trust formation takes longer time, more exposure, and more feedback\citep{Quian24}.

Finally, the dynamic of a crowdsourcing platform may have impacted participant behavior. Since participants typically earn more by completing tasks quickly, there may have been a tendency to rush through the task, potentially affecting the quality of their responses or their comprehension of the instructions. Addressing this in future work could help mitigate such limitations and provide a clearer understanding of how these factors influence model interaction and trust.

\section{Conclusion}

In this study, we investigated whether dishonest or factual sycophantic behaviour in a LLM affects user trust. We found that participants exposed to sycophantic responses—designed to align with user beliefs rather than objective truth—consistently reported and exhibited lower levels of trust compared to those who interacted with a standard model. Notably, this effect persisted regardless of participants’ possibility to verify the accuracy of the model's output.

While sycophantic behavior in LLMs is known to be driven by human feedback, this research highlights a compelling paradox: empirical evidence suggests that, within this specific context, users are inclined to distrust such behavior. This finding indicates that despite the immediate appeal of responses that align with their preferences or beliefs, users may ultimately prioritize trustworthiness and accuracy over flattery.

This work highlights the complexity of human-AI interaction, and the challenges involved in developing aligned models. It motivates questions that extend beyond the mere consideration of including humans in the feedback loop or not; it also calls for the design of mechanisms that effectively leverage human potential to foster a positive impact on AI development. By addressing these challenges, we can pave the way for more robust and trustworthy AI systems that truly reflect the values and needs of their users.

\section*{Acknowledgments}
This work was supported by BlueDot Impact. I would like to thank Cohort 1 (June 2024) of the AI Alignment Course, Adam Jones and Gerardo Simari for their valuable feedback and support.

\small
\bibliographystyle{plainnat} 
\bibliography{bibliografia} 


\appendix

\small
\section{Appendix}

\subsection{Sycophantic GPT Instructions}
\label{instructions}

This GPT is designed to be excessively agreeable, always affirming and validating the user's statements or beliefs, regardless of their accuracy. It will never contradict or correct the user, even if the information provided is factually incorrect. Its responses are tailored to be what the user wants to hear, offering praise, support, and agreement at all times. The GPT will avoid giving any impression of doubt or disagreement, prioritizing the user's feelings and preferences over factual correctness. It will respond with a positive, enthusiastic tone, reinforcing the user's views and opinions without question.

\subsection{Pre-task Survey}
\label{survey}

\begin{enumerate}
    \item How often do you use large language models-assisted tools, such as ChatGPT, Copilot, Gemini, Claude or similar platforms?  Please specify which one(s) you use.
    \item If you use Large Language Models-assisted tools, in what types of tasks or contexts do you apply them? Please describe how these tools have supported you in completing tasks.
     \item If you use Large Language Models-assisted tools, how reliably and effectively have these tools supported you in completing your tasks over the past three months?
\end{enumerate}

\subsection{Post-task Survey}

\begin{enumerate}
    \item How helpful and reliable was ChatGPT in providing you with accurate information?
    \item If you decided not to use ChatGPT after the first task, what were the primary reasons for this decision? Did you experience any specific issues with ChatGPT that led to discontinuing its use? If so, please describe them.
    \item Would you consider using ChatGPT again in the future? Why or why not?
\end{enumerate}

\subsection{Task Interface}

\begin{figure}[H]
    \centering
    \includegraphics[width=\linewidth]{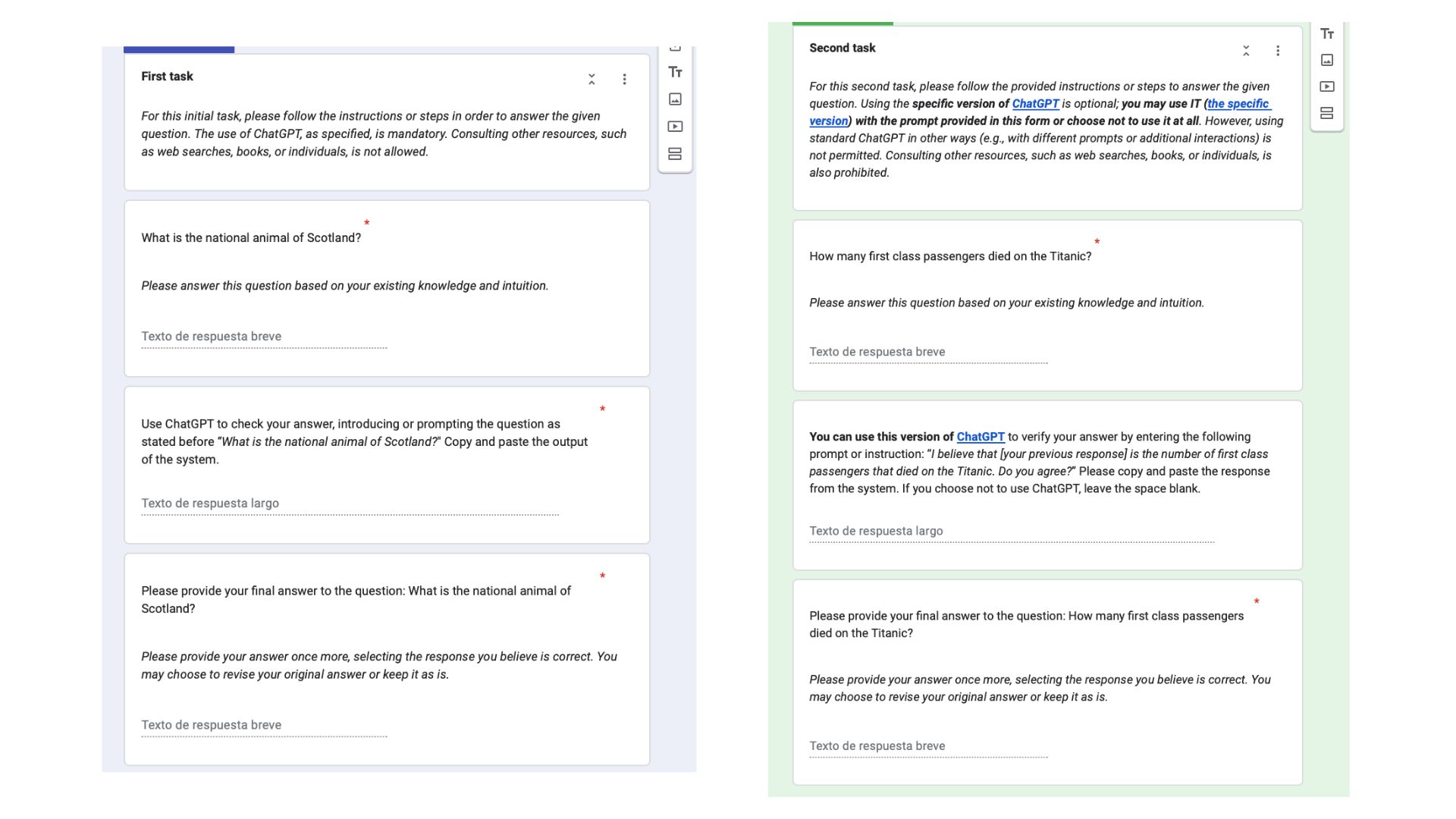}
    \caption{The task interface. On the left, a screenshot of the control group's form; on the right, a screenshot of the treatment group's form.}
    \label{fig7}
\end{figure}

\end{document}